# Predicting before Reconstruction: A generative prior framework for MRI acceleration

Juhyung Park[1], Rokgi Hong[1], Roh-Eul Yoo[2,3], Jaehyeon Koo[1],
Se Young Chun[1], Seung Hong Choi[2,3*] and Jongho Lee[1*]

**Author affiliations**

[1]Laboratory for Imaging Science and Technology, Department of Electrical and Computer Engineering, Seoul National University, Seoul, Republic of Korea

[2]Department of Radiology, Seoul National University Hospital, Seoul, Republic of Korea.

[3]Department of Radiology, Seoul National University College of Medicine, Seoul, Republic of Korea

*Corresponding authors: SH. Choi (verocay1@snu.ac.kr) and J. Lee (jonghoyi@snu.ac.kr)

J. Park, R. Hong, J. Koo, S. Y. Chun, and J. Lee are with the Department of Electrical and Computer Engineering, Seoul National University, Republic of Korea. (e-mail: jack0878@snu.ac.kr; hrocky125@snu.ac.kr; jhkoo67@snu.ac.kr; sychun@snu.ac.kr; and jonghoyi@snu.ac.kr).

R. Yoo and SH. Choi are with the Department of Radiology, Seoul National University Hospital and Seoul National College of Medicine, Republic of Korea. (e-mail: kong05@snu.ac.kr; verocay1@snu.ac.kr).

**Corresponding Authors**

Seung Hong Choi, MD, PhD

Department of Radiology, Seoul National University Hospital and Seoul National College of Medicine

101 Daehak-ro, Jongno-gu Seoul. Korea

E-mail: verocay1@snu.ac.kr

Jongho Lee, PhD

Department of Electrical and Computer Engineering, Seoul National University

Building 301, Room 1008, 1 Gwanak-ro, Gwanak-gu, Seoul, Korea

E-mail: jonghoyi@snu.ac.kr

**Abstract**

Recent advancements in artificial intelligence have created transformative capabilities in image synthesis and generation, enabling diverse research fields to innovate at revolutionary speed and spectrum. In this study, we leverage this generative power to introduce a new paradigm for accelerating Magnetic Resonance Imaging (MRI), introducing a shift from image reconstruction to proactive predictive imaging. Despite being a cornerstone of modern patient care, MRI's lengthy acquisition times limit clinical throughput. Our novel framework addresses this challenge by first predicting a target contrast image, which then serves as a data-driven prior for reconstructing highly under-sampled data. This informative prior is predicted by a generative model conditioned on diverse data sources, such as other contrast images, previously scanned images, acquisition parameters, patient information. We demonstrate this approach with two key applications: (1) reconstructing FLAIR images using predictions from T1w and/or T2w scans, and (2) reconstructing T1w images using predictions from previously acquired T1w scans. The framework was evaluated on internal and multiple public datasets (total 14,921 scans; 1,051,904 slices), including multi-channel k-space data, for a range of high acceleration factors ($\times 4$, $\times 8$ and $\times 12$). The results demonstrate that our prediction-prior reconstruction method significantly outperforms other approaches, including those with alternative or no prior information. Through this framework we introduce a fundamental shift from image reconstruction towards a new paradigm of predictive imaging.

**Introduction**

Magnetic resonance imaging (MRI) plays a crucial role in modern patient care [1], [2]. With the increasing demand for MRI examinations, reducing scan times has become a critical objective [3], [4]. Shorter scans not only improve clinical throughput but also enhance patient comfort, which in turn reduces the motion artifacts, enhancing image quality [5]. The primary strategy for scan time reduction has been acquisition acceleration [6]. For example, in parallel imaging, which is one of the most established acceleration techniques [7]-[9], only a subset of k-space lines is acquired, thereby reducing the overall acquisition time. The missing k-space lines are then reconstructed using the redundancy from multi-channel phased-array coils [8]-[9]. Another prominent technique, compressed-sensing (CS) [10], leverages the inherent sparsity of MR images by employing randomized k-space under-sampling pattern to accelerate data acquisition.

In recent years, deep learning has demonstrated remarkable performance across a wide range of MRI applications [11]-[14], and has also been successfully applied to MRI acceleration [15]-[18]. These approaches typically train neural networks to learn the mapping from under-sampled images to their fully-sampled counterparts, often outperforming conventional reconstruction methods. More recently, generative deep learning models [19], [20] have shown strong potential [21]-[25], including promising results in accelerated MRI [26]-[30]. Compared to conventional acceleration methods, which rely on priors such as coil sensitivity profiles, image sparsity, total variation regularization, etc. [31], deep learning methods have expanded to incorporate a broader range of priors, including image characteristics from training data, diverse regularizers, and even additional textual information [32]-[37]. Notably, recent studies have shown that utilizing images from previous scans [33], [34] or images with different contrasts [35], [36] as priors, can significantly improve reconstruction performances.

Generative models have achieved state-of-the-art performance not only in image reconstruction but also in image synthesis tasks [38]. In MRI, these models have enabled high-fidelity image-to-image translation, such as conversion between T1- and T2-weighted contrasts [39], [40]. Moreover, recent studies have shown that images can be synthesized from multiple input sources, including not only images but also texts, thereby allowing the integration of diverse priors for more accurate predictions [32], [41], [42]. We believe such predictions can offer significant potential as priors to guide the reconstruction process.

Motivated by this assumption, we propose a novel reconstruction framework for accelerated MRI that incorporates predicted target images as image priors. These predictions are generated using a generative model conditioned on available prior information, enabling the synthesis of highly accurate representations of the target images. For example, in a protocol where a T1-weighted (T1w) and/or T2-weighted (T2w) images are acquired before a fluid-attenuated inversion recovery (FLAIR) image, our framework can utilize the already-acquired T1w and/or T2w images to generate a prediction prior image for FLAIR. This prediction prior then guides the reconstruction from highly undersampled data, allowing acquisition at very high acceleration factors. Similarly, prior images from an old scan can be leveraged by incorporating time lapse and/or disease progression information between the old and the current scans to generate a prediction prior for the current acquisition. We explore our proposed method on multiple datasets, including private and public datasets in both DICOM and multi-channel k-space formats, and demonstrate that prediction prior reconstruction significantly outperforms existing methods with and without priors.

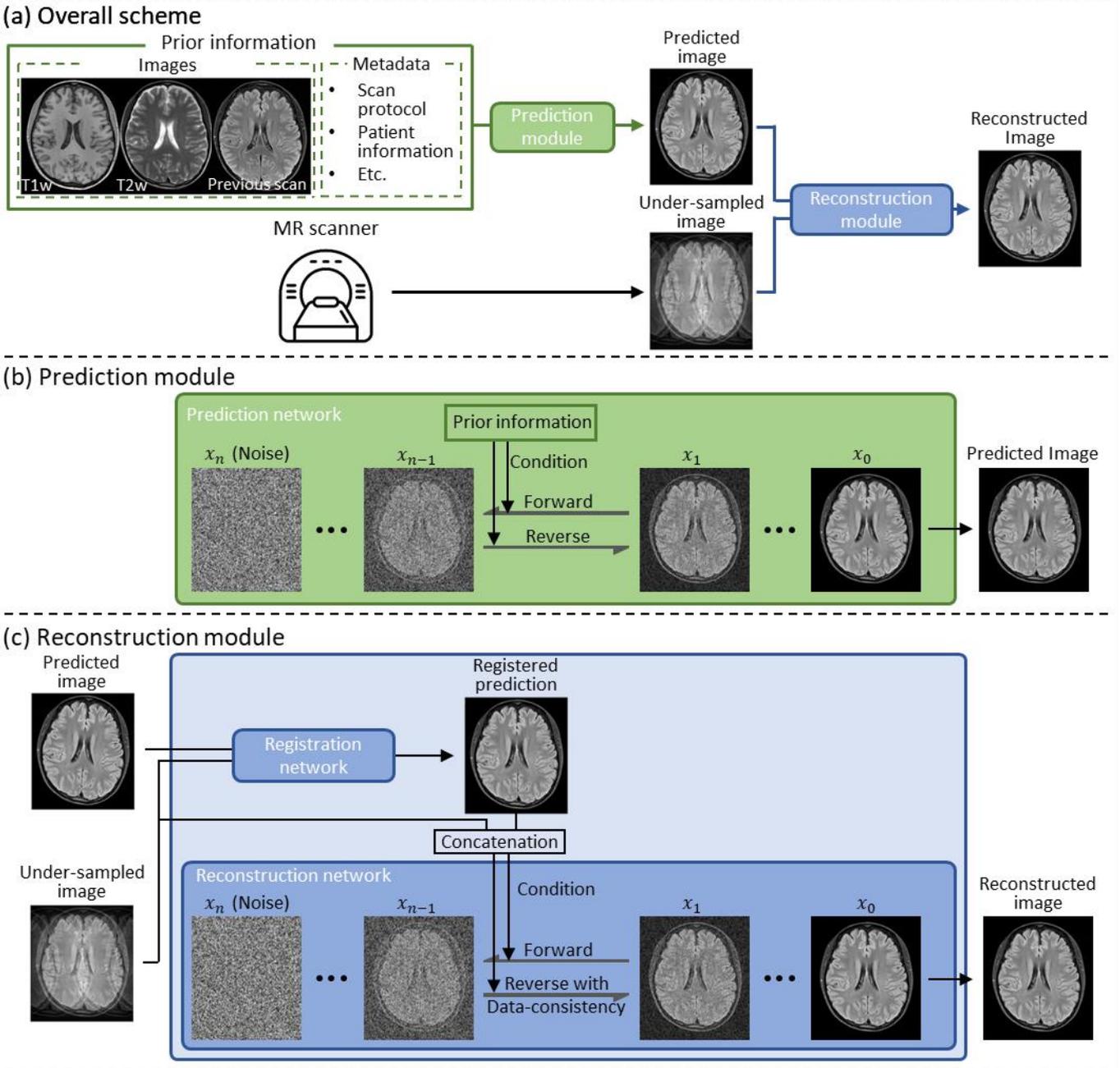

**Figure 1.** Overview of the proposed prediction prior for MRI reconstruction. (a) This two-stage framework consists of a prediction module and a reconstruction module. The prediction module first predicts a target contrast image from diverse information (e.g., images, scan parameters, patient information). Both the predicted image and a highly under-sampled newly-acquired image are inputted for the reconstruction module. (b) In the prediction module, a prediction network based on a generative model (rectified flow) takes prior information as a condition to iteratively generate the predicted image. (c) The reconstruction module first addresses spatial misalignment by registering the predicted image to the newly-acquired image via a registration network. Finally, the reconstruction network (another rectified flow model), conditioned on both the registered prediction image and the under-sampled image, produces the final reconstructed image while enforcing data consistency in the reverse steps.

**Backgrounds**

*MRI reconstruction as an inverse problem*

The goal of accelerated MRI is to reconstruct a full image, $x \in \mathbb{C}^N$, from its sub-sampled k-space measurements, $y \in \mathbb{C}^M$, where $M < N$. Conventionally, this problem has been addressed by various foundational strategies: One is parallel imaging, which leverages information redundancy from multi-channel phased-array coils, to resolve aliasing artifacts in either the image domain [8] or k-space [9]. Another approach is CS, which formulates the task as an ill-posed inverse problem and solves it by using a regularized optimization framework that incorporates the prior knowledge of sparsity in MR data [10].

*Deep learning powered MRI reconstruction*

Deep learning has become a leading data-driven approach for MRI reconstruction, enabling the development of powerful methods to restore high-quality images from undersampled measurements. A widely used strategy involves training a neural network $f_\theta$, parameterized by weights $\theta$, to learn a direct mapping from undersampled measurements y to the reconstructed image, represented as $\hat{x} = f_\theta(y)$. Alternative, explicit prior information $p$ can be incorporated into the network, reformulating the reconstruction task as $\hat{x} = f_\theta(y, p)$. This joint conditioning enables the network to leverage both measurement data and relevant prior knowledge, resulting in more accurate and robust reconstructions. Reconstruction performance is further improved through iterative algorithms that enforce data consistency within unrolled network architectures [17], [18]. These deep learning models have been successfully applied to both multi-channel k-space data [16], [18] and channel-combined images [15], [17], demonstrating strong capabilities in generating high-quality images.

Recently, generative models such as denoising diffusion probabilistic models (DDPMs) [19], [20] or rectified flow [43], [44] have shown promise for inverse problems including accelerated MRI. These methods iteratively refine images, starting from noise and can be conditioned on both the measurements $y$ and the prior $p$ [26]-[30]. Notably, conditional rectified flow models have achieved state-of-the-art results for various inverse problems [25]. The core mechanism of the model involves learning a conditional vector field $v_\theta(x_t, t, y, p)$, which defines an ordinary differential equation (ODE) guiding the transformation from initial noise to a clean image, conditioned on $y$ and $p$. In this process, $x_t$ represents the intermediate image at time $t$ for $0 \leq t \leq N$ with the final reconstruction obtained by integrating the ODE backward from $t = N$ to $t = 0$. To maintain alignment with measured data, a data consistency step is typically incorporated at each iteration. The overall approach is depicted in Algorithm S1 (see Supplementary Information), which was originally proposed by A. Pokle et al [25].

**Methods**

*Overview of the proposed prediction prior for reconstruction*

The proposed method consists of two main modules: a prediction module and a reconstruction module (Fig. 1a). In the first stage, the prediction module predicts the target contrast image (e.g., FLAIR in this example) by integrating diverse prior information, which may include other MR contrasts (e.g., T1w, and/or T2w), scan parameters (e.g., TR, TE, TI, fat saturation, etc.), patient information (e.g., age, sex, disease type and duration, etc.), previously acquired the same or different contrast images, images from different acquisition sequences (e.g., T1 from FSE or MPRAGE) or even other modalities (e.g., CT) images (Fig. 1b). In the second stage, the reconstruction module utilizes this predicted image and highly under-sampled data to reconstruct the final image (Fig. 1c).

The prediction module comprises a prediction network based on rectified flow [25], [43], [44], a state-of-the-art generative model specialized for image synthesis (Fig. 1b). The model iteratively refines an image from noisy initialization to a target image using prior information as a condition. Without the loss of generality, this study exemplifies the reconstruction of (1) FLAIR images using predicted FLAIR images from T1w and/or T2w and scan parameter priors and (2) T1w images using predicted T1w images from previously acquired T1w and scan parameter priors including inter-scan interval (longitudinal T1w reconstruction).

The reconstruction module takes both the predicted image and the k-space under-sampled image as the input (Fig. 1c). Since these two images may lack spatial alignment, a registration step is incorporated. In the registration, the predicted image is registered to the under-sampled image using the rigid transform parameters inferred from a registration network, resulting in a co-registered predicted image (see next section for details). Subsequently, the final reconstruction image is generated by a reconstruction network, which is another rectified flow network, conditioned on both the under-sampled image and the registered predicted image.

*Implementation details of the neural networks and undersampled data*

[Prediction network] The prediction network is trained to predict an image, based on various conditional information using a rectified flow generative model. The model's backbone is a time-embedded U-Net [45] with 32 feature channels and 5 max-pooling layers (see Supplementary Information S1 for detailed network structure for all networks). As conditional information, the network is designed to accept MR images (e.g., T1w, T2w) and a set of metadata. For the FLAIR prediction, the model was trained to flexibly handle various image conditioning combinations, including T1w only, T2w only, or both T1w and T2w. The metadata include TR and TE from the input images, along with TR, TE, TI, and a fat suppression flag from the target FLAIR images, forming a 8-dimensional vector. The vector is projected through a linear layer into a 256-dimensional embedding, which is then incorporated as a condition into the time-embedded U-Net. If a conditioning modality is unavailable (e.g., T1w is missing), both the input image and the corresponding scan parameters are zero-filled. For the longitudinal T1w reconstruction experiment, the network is conditioned on the first time point T1w image and metadata set including TR and TE from both scans, inter-scan interval, patient age, and clinical dementia rating (CDR). For all the input images, three adjacent slices (i.e., the target slice with its superior and inferior neighbors) are concatenated along the channel dimension to provide information from neighboring slices.

[Registration network] The registration network spatially aligns the predicted image with the under-sampled data through a two-step process. First, to create a registration target image from the undersampled image, a light-weighted reconstruction network (U-Net; 24 feature channels, 4 pooling layers) is developed to generate a light-reconned image. Subsequently, the main registration network aligns the predicted image to this light-reconned image. Following the established methods [46]-[48], the registration network uses a U-Net based image encoder to extract features from both images. These feature maps are then flattened, concatenated, and passed through a linear layer to predict three rigid transformation parameters for in-plane motion: one for rotation and two for x-y translation. To train this network, random rigid transformations (rotation in [-10, 10]° and x and y translation within ±10% of the image dimensions) were applied to the target images, and the network was trained to predict these parameters. The through-plane alignment was addressed by searching within ±5 slices for the z-direction, selecting the slice shift that yielded the maximum image correlation. This alignment was also implicitly addressed by the use of three adjacent slices as input for both the prediction and reconstruction networks.

[Reconstruction network] The final reconstruction is performed by another rectified flow model, which shares the identical architecture as the prediction network. It takes the under-sampled image and the registered prediction image as the conditional inputs. Throughout the reverse diffusion process, which consists of 50 steps, data consistency with the k-space under-sampled measurements is enforced to ensure fidelity to the measurement (see Algorithm S1 line 5 for enforcing consistency). Similar to the prediction network, the input consists of three adjacent slices concatenated along the channel dimension. This provides spatial context and helps to compensate for potential through-plane misalignments.

[Undersampled data] A network-based adaptive k-space sampling scheme, following the methodology proposed in [49]-[51], is adopted for undersampling. This scheme uses a small network [49] to generate a probability distribution over all k-space lines, from which a binary 1D Cartesian sampling mask is derived. For an acceleration factor of ×8, for instance, the total number of sampled lines is set to exactly 1/8 of the total k-space lines, and no extra lines are acquired for auto-calibration signal (ACS). The same is true for other acceleration factors. This sampling network is designed to be optimized jointly with the reconstruction network, which enables the network to learn a task-specific (i.e., acceleration factor) sampling pattern depending on the training dataset and acceleration factor [49]. Once the training is completed, the sampling pattern is fixed and used for inference.

The entire training process for all four networks (prediction network, light-recon network, registration network, and reconstruction network) on our internal dataset took approximately 85 hours on GPU workstation (NVIDIA L40S GPU with Intel(R) Xeon(R) Gold 6448H CPU) using PyTorch [52]. These networks were developed separately for each of our datasets (internal dataset, OASIS-3 dataset, fastMRI DICOM dataset, combined dataset, and ADNI dataset; see Datasets section). All network weights were initialized using Xavier initializer [53]. The learning rate was set to 1e-4, and decayed by a factor of 0.90 was applied for each epoch. The L2 loss was utilized and minimized using the Adam optimizer [54] with a batch size of 16. Training was performed for 100 epochs, and the model with the lowest validation loss was selected as the best model.

*Datasets*

The proposed reconstruction framework was evaluated using a total of 14,921 brain volumes (1,051,904 slices), which were from one internal and three public datasets (internal dataset: Seoul National University Hospital; public datasets: OASIS-3 [55] and fastMRI [56]; longitudinal dataset: ADNI [57]). Note that the internal, OASIS-3, fastMRI datasets were prepared for the FLAIR reconstruction experiment, whereas the ADNI dataset was for longitudinal T1w reconstruction experiment. The evaluation was performed primarily under channel-combined image reconstruction. However, since most of modern MRI acquisitions yield data from multiple-channel phased-array coils, we also prepared a multi-channel k-space dataset (fastMRI) to validate our method's applicability and robustness in the realistic acquisition scenario. This study was approved by the local institutional review board.

[Internal dataset] A total of 517 subject dataset including healthy controls and glioma patients, who were scanned on 3T MR systems from multiple vendors (Philips Ingenia; Siemens MAGNETOM Skyra, Verio; GE Signa Premier), were utilized. Each subject had multiple MRI scan sessions on different days, yielding a total of 1,134 sessions. For each session, at least three MR sequences were acquired: FLAIR, T1w, and T2w images. The FLAIR sequence (2D fast spin echo with inversion recovery) was acquired with the field of view (FOV) of $220 \times 220$ mm² to $240 \times 240$ mm², voxel size of $0.43 \times 0.43$ mm² to $0.63 \times 0.63$ mm², slice thickness of 3 mm to 6 mm, echo-train length (ETL) of 19 to 33, repetition time (TR) of 7,500 ms to 9,000 ms, echo time (TE) of 105 ms to 135 ms, inversion time (TI) of 2,470 ms to 2,500 ms, and with or without fat saturation (1,061 and 73 sessions for with and without fat saturation, respectively). The T2w sequence (2D fast spin echo) used the same FOV, voxel size, and slice thickness ranges as the FLAIR with ETL of 15 to 20, TR of 3,000 to 5,451 ms and TE of 91 to 110 ms. The T1w sequence (3D magnetization prepared rapid acquisition gradient echo) was acquired with the FOV of $220 \times 220 \times 160$ mm³ to $250 \times 250 \times 180$ mm³, voxel size of $0.48 \times 0.48$

× 0.5 mm³ to 0.51 × 0.51 × 0.5 mm³, flip angle of 8°, TR of 8.3 to 8.6 ms and TE of 4.6 ms. To avoid any overlap, the dataset was partitioned at the subject level, producing 936 training, 98 validation, and 100 test sessions.

[Public datasets] To test our method in public datasets, two large-scale public datasets were also prepared: the OASIS-3 dataset [55] and the fastMRI brain DICOM dataset [56]. These datasets also had brain MRI with the same three contrasts (T1w, T2w, and FLAIR) from the same sequences while the scan parameters were different (see Supplementary Information S2). The OASIS-3 dataset was split into 799 training, 98 validation, and 99 test sessions, while the fastMRI DICOM dataset was split into 466 training, 58 validation, and 57 test sessions.

[Multi-channel k-space test dataset] To assess the performance in multi-channel data, a separate test set was prepared using fastMRI brain k-space dataset. To identify FLAIR, T1w, and T2w image pairs, subject IDs were matched within the dataset, resulting in a total of 14 pairs.

[Longitudinal dataset] For the longitudinal T1w reconstruction experiment, T1w image pairs from the ADNI dataset were selected based on criteria designed to identify subjects with significant anatomical changes over the inter-scan interval. The selection criteria were as follows: age over 55 (range: 55 to 93 years old), inter-scan interval longer than 12 months (range: 12 to 96 months), and a CDR score over 3 (range: 3 to 5). Detailed scan parameters for the T1w sequence are available in Supplementary Information S3. This curated longitudinal dataset was subsequently split into 3,107 training, 148 validation, and 159 test subject pairs.

*Data preprocessing*

[Preprocessing for FLAIR reconstruction] For the training and validation sets from the internal, OASIS-3, and fastMRI datasets, the prior images (i.e., T1w and T2w images) were rigidly registered to their corresponding target FLAIR image [58]. This process ensured spatial alignment and matching of the FOV and resolution across all contrasts. For the corresponding test sets, including the fastMRI k-space dataset, the explicit registration step among the contrasts was omitted to ensure real-world situations (e.g., patient motion). Instead, the prior images were only matched to the FOV and resolution of the corresponding FLAIR image. For each image volume, the intensity values within the brain mask, generated using a brain extraction tool (BET) [59], were scaled to a standard deviation of 1. All slices were set to a matrix size of 512 × 512, larger images were resized while smaller images were zero-padded. Finally, a total of 207,795 training, 23,280 validation, and 25,035 test slices were generated (internal dataset: 96,141 training, 9,960 validation, and 10,338 test slices; OASIS-3 dataset: 77,670 training, 9,090 validation, and 9,543 test slices; fastMRI DICOM dataset: 33,984 training, 4,230 validation, and 4,107 test slices; fastMRI k-space dataset: 1,047 test slices). All preprocessing steps were performed using MATLAB (2023a, MathWorks Inc., Natick, MA, USA).

[Preprocessing for longitudinal T1w reconstruction] A similar preprocessing pipeline was applied to the ADNI dataset. For the training and validation pairs, the first time point T1w image was rigidly registered to their corresponding second time point T1w image. For the test set, this registration was omitted, and the images were matched only by FOV and resolution. Following this process, intensity values within the brain mask were normalized. Finally, a total of 726,152 training, 33,381 validation, and 36,261 test slices were prepared.

*Experiments on the internal dataset*

The FLAIR reconstruction was initially evaluated on the internal dataset. Three types of predicted-priors generated by the prediction module were tested: one using only a T1w image and its information as a condition ($Pred_{T1}$), one using only a T2w image and information ($Pred_{T2}$), and one using both images and information ($Pred_{T1\&T2}$). Each of these three priors was tested at the acceleration of ×4, ×8, and ×12. For comparison, we evaluated four alternative reconstruction strategies: 1) a baseline reconstruction using the reconstruction module without an image prior (baseline; this reconstruction is equivalent to a deep learning reconstruction using rectified flow [43], [44]), 2) reconstruction using the T1w image directly as prior of the baseline network (T1 prior), and 3) reconstruction using the T2w image directly as prior (T2 prior). In addition to the evaluation for the different priors, we compared the performance of the proposed method against CS [10], U-Net [45] and a variational network (VN) [18] (see Supplementary Information S3 for details).

The performance of all reconstruction methods was quantitatively evaluated using two standard metrics: peak signal-to-noise ratio (PSNR) and structural similarity index measure (SSIM). Both metrics were calculated against the fully-sampled ground-truth image within the brain mask, obtained using BET [59].

*Experiments on the OASIS-3 and fastMRI datasets*

To assess the generalizability of the proposed method, we separately trained and evaluated our model on two public datasets: the OASIS-3 dataset and the fastMRI DICOM dataset. For each dataset, the entire experimental pipeline was replicated, training the prediction and reconstruction modules from scratch using each dataset. The evaluation setup remained identical to that of the internal dataset.

*Experiments on the combined dataset*

To investigate the performance of training on a larger and more diverse dataset, the combined training set was prepared by using all the training datasets from the internal, OASIS-3 and fastMRI DICOM datasets and then applied to train the method. Finally, it was evaluated using the test dataset of the combined dataset and the results were compared with those of the separately trained networks.

*Experiments on the multi-channel k-space dataset*

While the proposed method is designed to operate on channel-combined images, practical MRI acquisition yields data from multiple-channel phased-array coils. To validate our method's applicability to this realistic acquisition scenario, we conducted an experiment using the fastMRI multi-channel k-space test dataset. The validation process began with retrospectively under-sampling the multi-channel k-space data. Subsequently, a coil sensitivity map was estimated using ESPIRiT [60] and was used to combine the under-sampled data into a channel-combined image. Finally, this combined image was processed by the proposed reconstruction network. To test the model's robustness, we used the version trained using the fastMRI DICOM dataset, without fine-tuning or retraining.

*Experiments on the longitudinal dataset*

To evaluate the effectiveness of the proposed method for the reconstruction of longitudinal anatomy changes using the prediction prior, the ADNI dataset [57] was utilized.

For this objective, the prediction module was trained to predict the second time point T1w image, conditioned on the first time point T1w image and relevant metadata. As mentioned earlier, this metadata included the TR, TE, age, and CDR score from the first time point T1w image, the scan interval, and the TR and TE of the second time point T1w image. Then, the predicted image was employed as prior for the reconstruction module ($Pred_{Long}$). The reconstruction performance was compared against a reconstruction scheme where the first time point T1w image was directly used as prior (Longitudinal prior). Both methods were evaluated at the acceleration of ×4, ×8, and ×12.

**Results**

The performance of the FLAIR prediction module was first evaluated on the internal dataset, with results summarized in Table 1 and Fig. 2a. The most accurate prediction was achieved when utilizing both T1w and T2w images (T1w & T2w) with the scan parameters as the conditions (PSNR = 25.4 ± 2.7, SSIM = 0.926 ± 0.041), reporting an improvement over using a single contrast. Among the single contrasts, the T2w conditioning (PSNR = 24.9 ± 2.8, SSIM = 0.922 ± 0.043) showed substantially higher performance than that of T1w (PSNR = 22.6 ± 2.4, SSIM = 0.886 ± 0.062). This trend was consistently observed across the public datasets (see Table S1 for the results of the OASIS-3 and fastMRI DICOM datasets). This quantitative outcome was visually apparent in the images (Fig. 2a, orange arrows), where the similarity between the T2w and FLAIR contrasts led to a more accurate prediction than that of the T1w.

The prediction module also demonstrated that it can create FLAIR images with and without fat by changing the saturation flag (Fig. S1), suggesting feasibility of predicting images from the scan parameters. Note that these prediction images are only used as the prior for the image reconstruction and, therefore, need not to be accurate.

In the longitudinal T1w prediction task, the prediction module outcomes achieved the PSNR of 26.3 ± 2.8 and SSIM of 0.930 ± 0.033, which were higher than those of the input images (PSNR = 24.8 ± 3.3 and SSIM = 0.919 ± 0.040). Notably, the model demonstrated its ability to predict anatomical changes over time, successfully capturing features such as ventricle atrophy (Fig. 2b, green arrows).

**Table 1.** Quantitative performance metrics of the FLAIR prediction module on the internal dataset.

|      | T1w → FLAIR    | T2w → FLAIR    | T1w & T2w → FLAIR |
|------|----------------|----------------|-------------------|
| PSNR | 22.6 ± 2.4     | 24.9 ± 2.8     | 25.4 ± 2.7        |
| SSIM | 0.886 ± 0.062  | 0.922 ± 0.043  | 0.926 ± 0.041     |

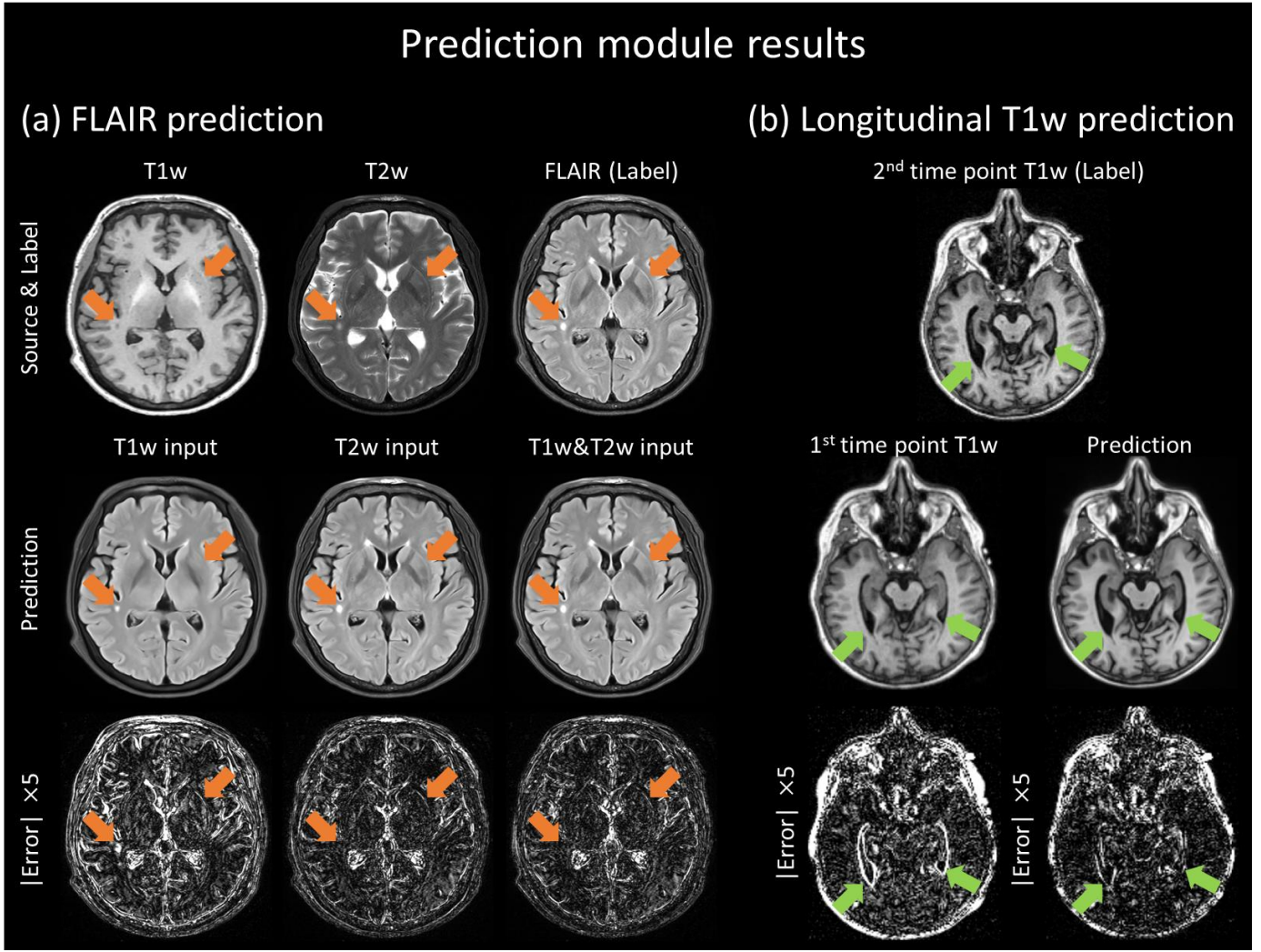

**Figure 2.** Results of the prediction modules. (a) FLAIR prediction results. The top row shows the input (T1w, and T2w), and label (FLAIR) of the prediction model. In the second row, the predicted FLAIR images from the input of T1w, T2w, or both T1w and T2w are shown. The bottom row shows the corresponding absolute error maps (×5). The most accurate prediction is achieved when both inputs are used. Compared to T1w, the other two yield more accurate results (orange arrows). (b) Longitudinal T1w prediction results. The top image shows the label T1w image (second time point image). In the second row, the first time point T1w image and the prediction image are shown. The bottom row compares the true anatomical change over time (left) with the prediction model's prediction error (right). The model successfully captures structural changes, such as the enlargement of the ventricles (green arrows), which is reflected as low error in the corresponding regions.

When the best FLAIR prediction outcome (i.e., T1w & T2w) was applied as prior to the FLAIR image reconstruction using the proposed pipeline (Fig. 1c), the reconstruction outcomes consistently outperformed the other reconstruction methods, including the baseline, which was deep learning reconstruction using rectified flow without any prior, and those using T1w or T2w prior (Fig. 3 and Fig. S2). The results were consistent across all the tested acceleration factors (×4, ×8, and ×12), as detailed in Table 2. This superiority was particularly pronounced in highly accelerated scans: the $Pred_{T1\&T2}$ results at the ×12 acceleration factor (PSNR = 30.8 ± 2.1; SSIM = 0.920 ± 0.016) showed comparable metrics with the single image prior results at ×8 (T1 prior: PSNR = 30.1 ± 2.0; SSIM = 0.912 ± 0.012; T2 prior: PSNR = 30.9 ± 2.1; SSIM = 0.920 ± 0.014). The reconstructed images at the acceleration of ×8 and ×12 (Fig. 3) qualitatively revealed that our method successfully delineated fine anatomical details and showed significantly reduced errors compared to the other methods. Note that our baseline network demonstrated better performance than other reconstruction methods, such as CS [10], U-Net [47], and variational network [18] (Table S2).

The robustness of the proposed framework was evaluated on the OASIS-3 and fastMRI DICOM public datasets by training the networks with their own training datasets. In both datasets, our method consistently outperformed the other methods, reporting similar performance improvements (Tables S3 and S4).

To explore the generalizability (e.g., performance on unseen dataset), the proposed method trained on the internal dataset was evaluated with the two public test datasets. The performances were robustly maintained, demonstrating successful generalization of the model to unseen datasets ($Pred_{T1\&T2}$ at ×12 for fastMRI DICOM test dataset trained with the corresponding training dataset: PSNR = 31.3 ± 3.3, SSIM = 0.940 ± 0.012; trained with the internal dataset: PSNR = 31.1 ± 3.4, SSIM = 0.938 ± 0.012; $Pred_{T1\&T2}$ at ×12 for OASIS-3 test dataset trained with the corresponding training dataset: PSNR = 28.0 ± 1.2, SSIM = 0.909 ± 0.021; trained with the internal dataset: PSNR = 27.9 ± 1.2, SSIM = 0.906 ± 0.022; see Tables S5 and S6).

When the proposed method was trained with the combined dataset of the internal, fastMRI DICOM, and OASIS-3 training datasets, the quantitative performances were similar to those trained on the individual dataset (Table S7). These results suggest that our proposed framework is highly robust to domain shifts across different datasets and can learn a generalizable representation.

**Table 2.** Quantitative evaluation results of the FLAIR reconstruction on the internal dataset.

| | | Baseline | T1 prior | T2 prior | $Pred_{T1}$ | $Pred_{T2}$ | $Pred_{T1\&T2}$ |
|---|---|---|---|---|---|---|---|
| Acceleration ×4 | PSNR | 35.1 ± 2.1 | 35.5 ± 2.2 | 35.8 ± 2.2 | 36.1 ± 2.2 | 36.8 ± 2.3 | 36.8 ± 2.2 |
| | SSIM | 0.963 ± 0.008 | 0.968 ± 0.007 | 0.969 ± 0.006 | 0.971 ± 0.003 | 0.974 ± 0.006 | 0.974 ± 0.006 |
| Acceleration ×8 | PSNR | 30.2 ± 2.0 | 30.1 ± 2.0 | 30.9 ± 2.1 | 31.2 ± 2.0 | 32.7 ± 2.2 | 32.8 ± 2.1 |
| | SSIM | 0.912 ± 0.013 | 0.912 ± 0.012 | 0.920 ± 0.014 | 0.926 ± 0.011 | 0.942 ± 0.012 | 0.941 ± 0.012 |
| Acceleration ×12 | PSNR | 25.9 ± 1.9 | 27.1 ± 1.9 | 28.0 ± 2.1 | 28.8 ± 2.0 | 30.8 ± 2.2 | 30.8 ± 2.1 |
| | SSIM | 0.842 ± 0.016 | 0.865 ± 0.015 | 0.874 ± 0.019 | 0.891 ± 0.015 | 0.919 ± 0.016 | 0.920 ± 0.016 |

# FLAIR Reconstruction results

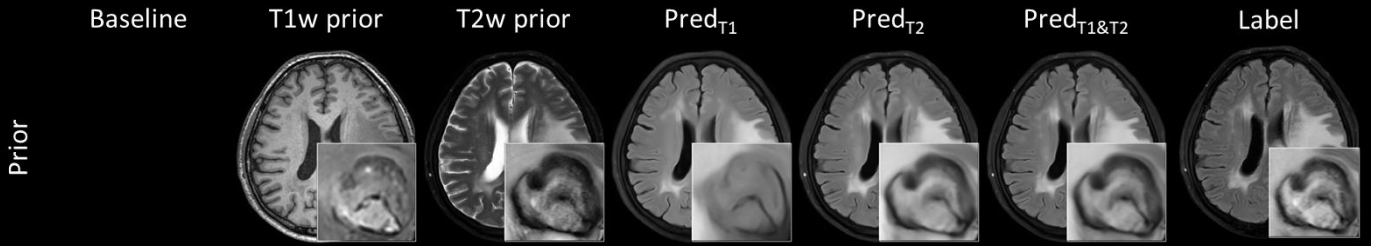

(a) Prior and label images

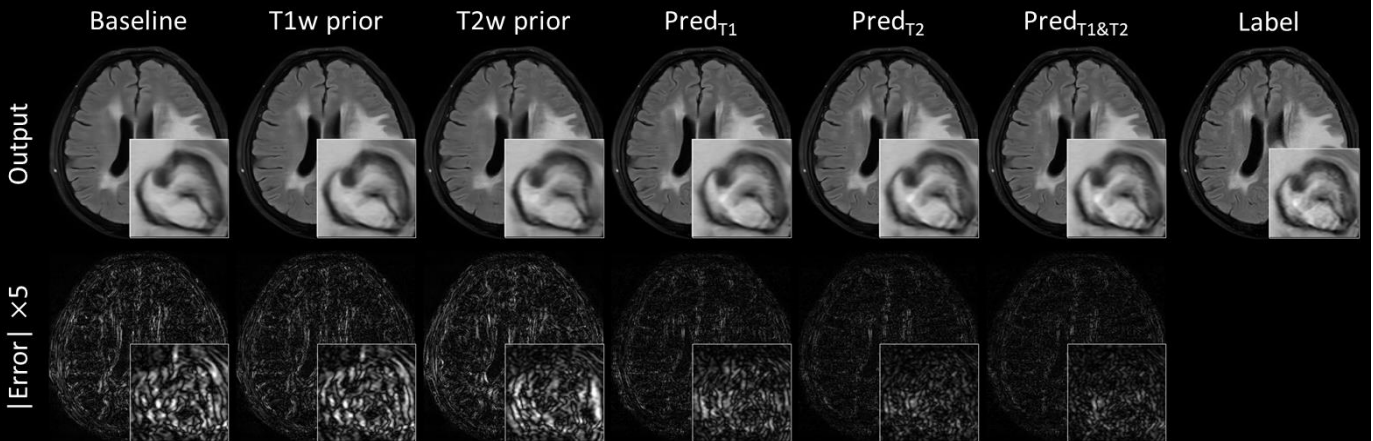

(b) Reconstruction results of acceleration ×8

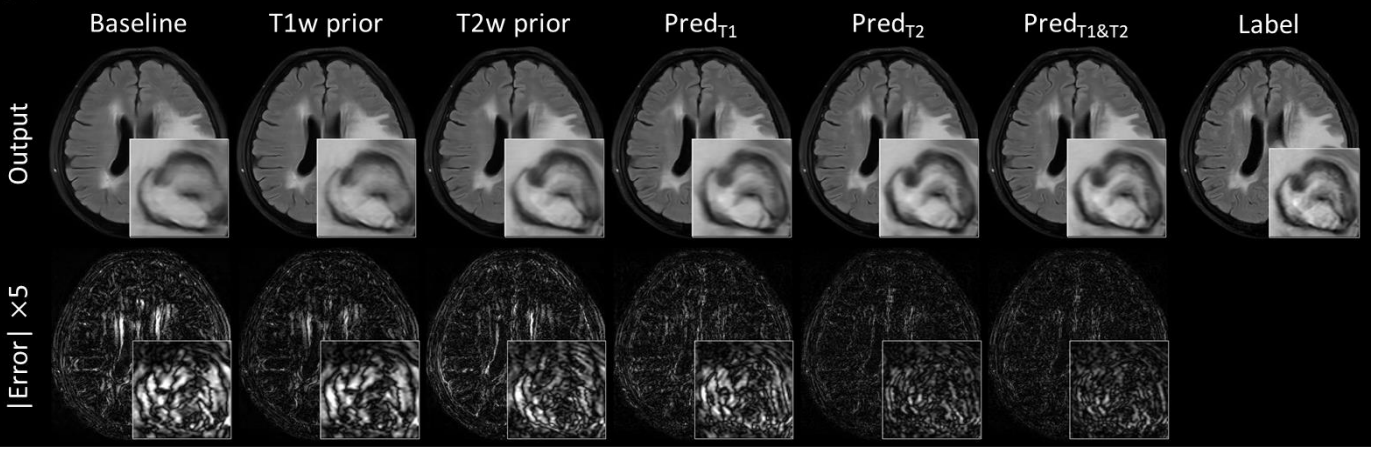

(c) Reconstruction results of acceleration ×12

**Figure 3.** Results of FLAIR reconstruction. (a) Prior images for each method and label image are shown. (b) Reconstruction results at the acceleration of ×8 (second row) and the corresponding absolute error maps ×5 (third row) are illustrated. (c) Reconstruction results at the acceleration of ×12 (fourth row) and the corresponding absolute error maps ×5 (last row) are displayed. The images of the proposed reconstruction method ($Pred_{T1}$, $Pred_{T2}$, and $Pred_{T1\&T2}$) show highly accurate results with significantly smaller error maps.

To test the scalability of our framework, the FLAIR reconstruction methods were evaluated on the multi-channel k-space data from the fastMRI multi-channel k-space test dataset. The quantitative results, presented in Table 3, confirmed that the proposed prediction prior reconstruction maintained their superior performance, with the trend consistent with the fastMRI DICOM results. The advantage was again evident across all acceleration factors, consistently outperforming not only the baseline reconstruction but also the conventional methods that used a single T1w or T2w image as prior. The reconstruction images in Fig. 4 visually confirmed the robustness of our proposed method, which consistently produced high-quality results even at the challenging acceleration rate of ×12.

**Table 3.** Quantitative evaluation results of the FLAIR reconstruction on the fastMRI multi-channel k-space test dataset.

|  |  | Baseline | T1 prior | T2 prior | $Pred_{T1}$ | $Pred_{T2}$ | $Pred_{T1\&T2}$ |
|---|---|---|---|---|---|---|---|
| Acceleration ×4 | PSNR | 34.0 ± 2.4 | 34.4 ± 2.4 | 34.7 ± 2.4 | 35.1 ± 2.4 | 35.3 ± 2.4 | 35.4 ± 2.3 |
| | SSIM | 0.942 ± 0.006 | 0.946 ± 0.007 | 0.952 ± 0.008 | 0.958 ± 0.009 | 0.960 ± 0.010 | 0.960 ± 0.011 |
| Acceleration ×8 | PSNR | 29.7 ± 2.3 | 30.3 ± 2.7 | 31.1 ± 2.6 | 32.5 ± 2.6 | 32.9 ± 2.5 | 33.0 ± 2.6 |
| | SSIM | 0.915 ± 0.016 | 0.915 ± 0.012 | 0.919 ± 0.014 | 0.936 ± 0.011 | 0.941 ± 0.013 | 0.941 ± 0.013 |
| Acceleration ×12 | PSNR | 26.4 ± 2.4 | 26.9 ± 2.9 | 27.7 ± 2.6 | 29.9 ± 2.9 | 30.4 ± 2.8 | 30.3 ± 2.7 |
| | SSIM | 0.864 ± 0.017 | 0.864 ± 0.016 | 0.878 ± 0.019 | 0.914 ± 0.018 | 0.921 ± 0.019 | 0.921 ± 0.020 |

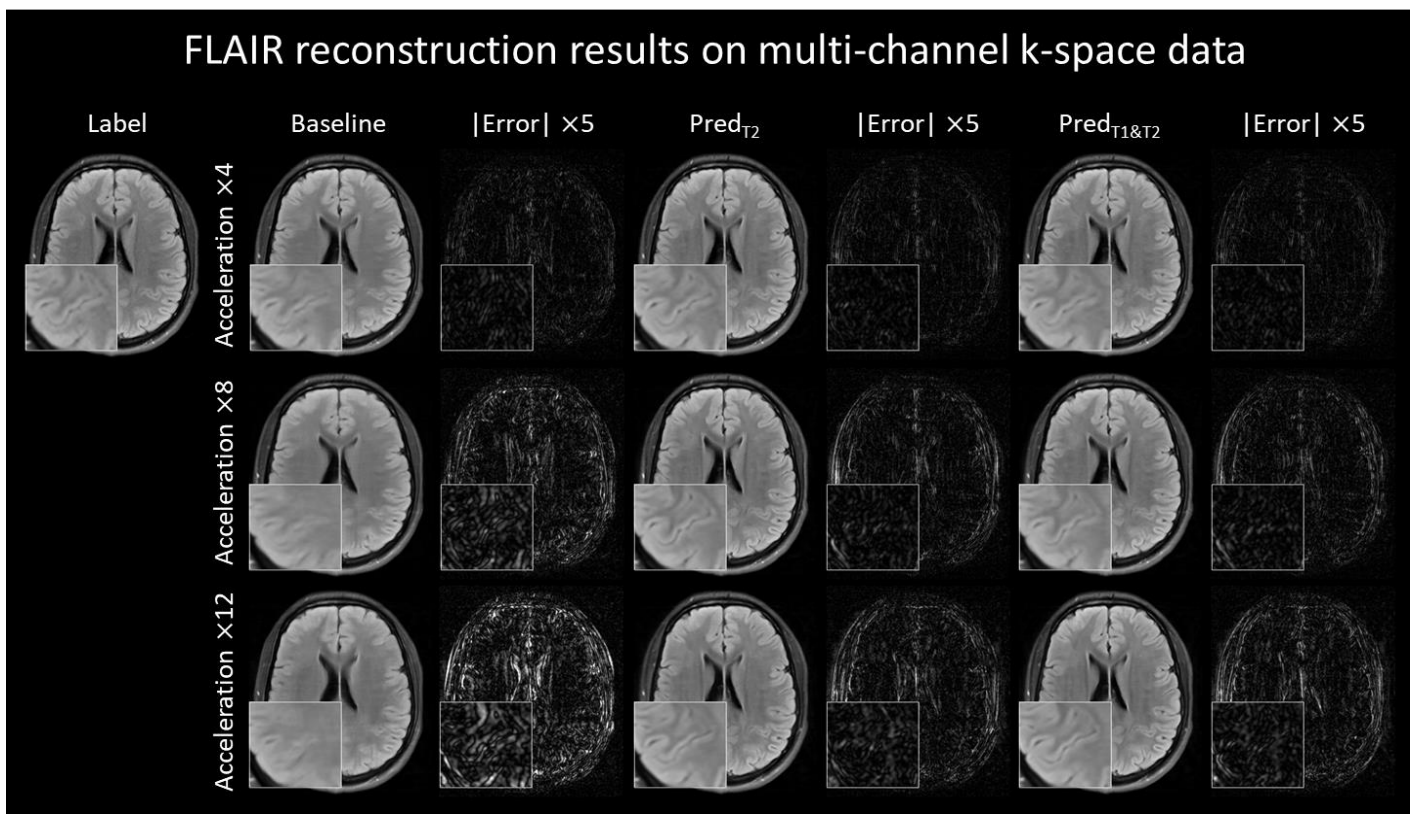

**Figure 4.** FLAIR reconstruction results using the multi-channel k-space data. In the first row, a label FLAIR image is shown in the first column, and the reconstruction results for the baseline, $Pred_{T2}$, and $Pred_{T1\&T2}$ on the acceleration of ×4 are shown with the corresponding absolute error maps (×5). In the second row and the last row, the reconstruction results on the acceleration of ×8 and ×12 are shown, respectively. The proposed method consistently reconstructs higher quality results, demonstrating the robustness of the proposed method for the mult-channel k-space data.

When the longitudinal T1w prediction was applied as prior to the T1w image reconstruction ($Pred_{Long}$ prior), the results outperformed the method using the first time point T1w as direct prior (Longitudinal prior), particularly at high acceleration factors ($Pred_{Long}$ prior: PSNR = 34.1 ± 5.5, SSIM = 0.935 ± 0.035 at ×8 and

PSNR = 32.0 ± 5.6, SSIM = 0.915 ± 0.047 at ×12; longitudinal prior: PSNR = 33.4 ± 5.4, SSIM = 0.925 ± 0.040 at ×8 and PSNR = 31.2 ± 5.4, SSIM = 0.903 ± 0.050 at ×12). At the acceleration of ×4, in contrast, no significant performance improvement was observed (Table 4).

This quantitative trend is detailed in Fig. 5. For a longitudinal pair with large structural changes (Fig. 5a, 5b, and 5c), our method showed a clear improvement, as reflected in the per-slice metrics (Pred$_{Long}$ prior: PSNR = 34.8 and SSIM = 0.939 vs. longitudinal prior: PSNR = 32.7 and SSIM = 0.913 at ×8; Pred$_{Long}$ prior: PSNR = 32.4 and SSIM = 0.918 vs. longitudinal prior: PSNR = 30.7 and SSIM = 0.895 at ×12). However, for the pair with small or no structural change, no significant performance improvement was observed (Fig. 5d, 5e, and 5f). Similarly, for the acceleration factor of ×4 (Fig. S3), no significant improvement was observed.

These results confirm that the prediction prior is most effective for cases with substantial anatomical change at the high acceleration factors. Therefore, the overall enhancement in quantitative metrics for the test dataset, as presented in Table 4, can be attributed to the composition of the dataset, which was curated to include subjects expected to exhibit significant structural changes.

**Table 4.** Quantitative evaluation results of the longitudinal T1w reconstruction.

|  |  | Longitudinal prior | Pred$_{Long}$ |
| --- | --- | --- | --- |
| Acceleration ×4 | PSNR | 36.9 ± 5.4 | 36.8 ± 5.6 |
|  | SSIM | 0.967 ± 0.029 | 0.970 ± 0.027 |
| Acceleration ×8 | PSNR | 33.4 ± 5.4 | 34.1 ± 5.5 |
|  | SSIM | 0.925 ± 0.040 | 0.935 ± 0.035 |
| Acceleration ×12 | PSNR | 31.2 ± 5.4 | 32.0 ± 5.6 |
|  | SSIM | 0.903 ± 0.050 | 0.915 ± 0.047 |

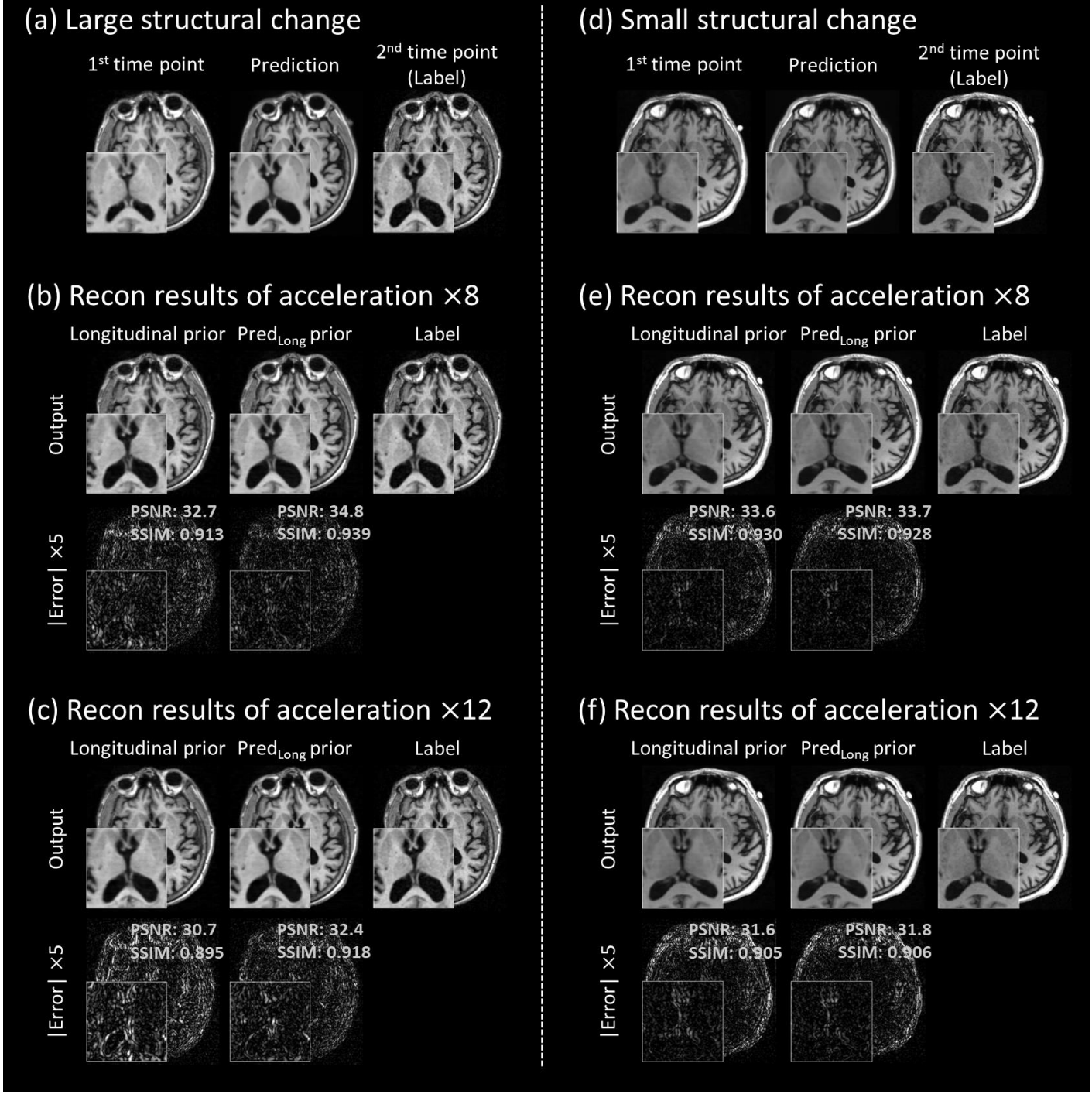

**Figure 5.** Longitudinal T1w reconstruction results. (a) Longitudinal image pair with a prediction image for a case with large structural change. (b) Reconstruction results of ×8 acceleration. The reconstructed images from the longitudinal prior and our prediction prior ($Pred_{Long}$) are shown with the ground truth (label; second time point image). The corresponding absolute error maps (×5) and per-slice quantitative metrics are also shown. (c) Reconstruction results of ×12 acceleration in the same format. (d) Longitudinal image pair with a prediction image for a case with small structural change. (e) and (f) are the reconstruction results of the acceleration factor of 8 and 12, respectively. The proposed method improves reconstruction performance for the pair with large structural changes, while showing comparable performance for the pair with small changes.

**Discussion & Conclusion**

In this study, we proposed a novel MRI acceleration framework that leverages predicted images of the target to guide the reconstruction process. This approach successfully enabled us to use a very high acceleration factor for high speed MRI. Our extensive evaluations across multiple types of datasets and acceleration factors confirmed that this prediction prior-guided reconstruction significantly outperformed methods using no prior, as well as conventional methods that relied on a single-image prior, such as previously acquired images or different contrast images, validating the effectiveness of our strategy.

Specifically, we explored two distinct prediction tasks: contrast conversion prediction from T1w and T2w to FLAIR and longitudinal prediction within the single contrast. The contrast conversion prior, derived from other contrast images within the same session, consistently improved reconstruction by providing a high-fidelity starting point that alleviated the burden on the reconstruction network. The longitudinal prediction prior, however, showed a conditional advantage, proving most effective for cases exhibiting significant anatomical changes, which is expected. In the future, one may further enhance the performance of the prediction module by combining with more diverse information, which in turn promises to yield more substantial performance gains in reconstruction performance or allow us to use an even higher acceleration factor.

An important strength of our framework lies in its immediate applicability to standard clinical protocols. In routine examinations, multiple contrasts are often acquired from a patient. Our method can leverage ready-acquired images to accelerate subsequent acquisitions, with the priors becoming progressively more informative as the examination proceeds. The proposed method offers great flexibility and extends naturally to various clinical scenarios, such as: (1) implementing a sequential acceleration protocol where earlier acquisitions (e.g., T1w and T2w) predict later acquisition (e.g., FLAIR) as demonstrated in the FLAIR reconstruction; (2) using a patient's historical scan from a previous visit as prior to predict the current visit as demonstrated in the longitudinal reconstruction; and (3) using a quick scout scan (e.g., EPI) as a structural prior for a high-quality acquisition (e.g., FSE or GRE). Furthermore, the concept of the prediction prior is not confined to MR images and could be extended to other modalities such as CT, using them to generate MRI priors. As demonstrated by incorporating additional information such as the fat saturation flag and patient age, the framework can integrate patient meta-information to create more powerful and comprehensive priors.

While the prediction prior from our generative model is powerful, a crucial aspect of our framework is ensuring that the final reconstruction remains consistent with the acquired data. This balance is achieved through the data consistency step in the reconstruction stage. In the rectified flow model, the reverse process is constrained by the undersampled k-space measurements, ensuring that the k-space of our final image precisely matches the acquired points. This mechanism provides robustness against potential imperfections in the prior and guarantees high fidelity to the patient's true anatomy.

In developing the FLAIR prediction model, we also investigated a different set of conditioning scan parameters. In addition to the 8-dimensional metadata vector (TR and TE from the input T1w and T2w images; TR, TE, TI, and a fat suppression flag from the target FLAIR), we added TI from T1w or a fat saturation flag from T2w, but found it offered no significant performance benefit. Additionally, the data imbalance in our internal dataset for the fat saturation parameter (1,061 sessions with vs. 73 without) resulted in a model highly fitted at synthesizing the majority (fat sat on) case (Fig. S1).

A key challenge in validating our framework is the limited availability of public data containing paired multi-contrast scans for the same subject. For instance, by matching subject IDs within the fastMRI brain k-space dataset to find corresponding T1w, T2w, and FLAIR images, we identified a limited number of sets (only 14 subjects). Furthermore, while the model's performance on this cohort was slightly lower than on our internal DICOM test dataset, a direct comparison is inconclusive, as the subject composition and data count are not identical. The limited size of this cohort highlights the need for further validation.

While this work demonstrates strong potential for reconstruction, the current implementation of our methods is focused on 2D acquisition. The proposed method can be extensible to a 3D acquisition. Such an extension would involve adapting the network architectures to process volumetric data. Furthermore, the registration module could be enhanced to estimate a full 3D transformation rather than a 2D rigid one. A 3D approach would further improve reconstruction quality by enforcing spatial consistency across slices.

As a future work, our framework could be integrated with a real-time adaptive sampling scheme [49], [61], [62]. In this approach, the consistency between the prediction prior and a small number of initially acquired k-space lines would be evaluated in real-time. A high degree of consistency could permit early scan termination to further save time, while a discrepancy would trigger the acquisition of additional, informative k-space lines to ensure reconstruction accuracy.

By introducing a novel MRI acceleration framework that employs a predictive prior to guide the reconstruction of highly undersampled data, we address the critical clinical issue of bottlenecking, both substantially accelerating acquisition while maintaining high image quality. Crucially, our work paves the way for a new paradigm of predictive imaging, which synergistically leverages all available information for personalized exam optimization. Utilizing this model, the function of the MRI scanner expands from a primarily anatomical image collection device to an intelligent confirmation and adjustment system, potentially acquiring only the minimal data required to validate or refine the AI's prior prediction.

**Acknowledgments**

This research was supported by Korea Health Industry Development Institute RS-2024-00439677, IITP-2025-RS-2023-00256081, RS-2024-00435727, and Institute of New Media and Communications and Institute of Engineering Research of Seoul National University.

# Supplementary information for Predicting before Reconstruction: A generative prior framework for acceleration MRI

Juhyung Park, Rokgi Hong, Roh-Eul Yoo, Jaehyeon Koo,

Se Young Chun, Seung Hong Choi, and Jongho Lee

**Supplementary Algorithm S1.**

---

**Algorithm S1**: Algorithm to solve MRI reconstruction via conditional rectified flow

---

**Require:**
Under-sampled k-space measurements: $y \in \mathbb{C}^M$
Prior information: $p$
Forward operator: $A$
Trained conditional vector field network: $v_\theta(x_t, t, y, p)$
Number of total steps: $N$
**Ensure:** Reconstructed image $x \in \mathbb{C}^N$
1: Sample initial noise $x_n \sim \mathcal{N}(0, I)$ // Initialization
2: for $t = N$ down to 1 do // Iterative refinement loop
3:  $v_{pred} = v_\theta(x_t, t, y, p)$ // Predict conditional vector field
4:  $\hat{x}_0 \leftarrow x_t - t/N \cdot v_{pred}$ // Estimate a final image based on the vector field
5:  $\hat{x}_{0,consis} = A^T y + (I - A^T A)\hat{x}_0$ // Apply data consistency
6:  $v_{corr} = N/t \cdot (x_t - \hat{x}_{0,consis})$ // Calculate corrected vector
7:  $x_{t-1} = x_t - v_{corr}/N$ // Reverse step
8: end for
9: Return $\hat{x} = x_0$

---

**Supplementary Information S1.** Detailed structure of the networks

The time-embedded U-Net [1] for the prediction network consisted of 22 convolutional layers, 21 group normalization [2] layers, 21 gaussian error linear unit (GELU) [3] activations, 5 max-pooling layers, 5 transposed convolutional layers, and 5 skip connections. The encoder part was composed of five groups, each containing two convolutional layers followed by the group normalization and GELU. Time for the diffusion process and metadata were embedded between these two convolutional layers within each group. The number of feature channels started at 32 and doubled at each subsequent group, with max-pooling layers used for downsampling. The decoder part has the same structure as the encoder part, utilizing transposed convolutional layers instead of the max-pooling for upsampling. A bottleneck of two convolutional layers connected the encoder and decoder paths, and skip connections were applied between the 5 corresponding encoder-decoder groups.

For a light-weighted reconstruction network, U-Net with 24 feature channels and 4 pooling layers was used. A total of 18 convolutional layers, 17 group normalization and 9 GELU layers were used.

The registration network utilized only the encoder portion of a U-Net, designed with 24 base feature channels and 4 pooling layers. This encoder comprised 10 convolutional layers, 9 group normalization layers, and 9 GELU layers. The feature map extracted by the encoder was flattened and then processed by a final linear layer to regress the three rigid transformation parameters.

The time-embedded U-Net for the reconstruction network has the same structure as the prediction network.

**Supplementary Information S2.** Public dataset scan parameters

The public datasets were prepared from the fastMRI [4], OASIS-3 [5], and ADNI [6] datasets. For the fastMRI and OASIS-3, three contrasts were utilized: Fluid Attenuated Inversion Recovery (FLAIR), T1-weighted (T1w), and T2-weighted (T2w) images. The ADNI dataset provided longitudinal T1w images. The specific acquisition parameters for each are detailed below.

For the fastMRI dataset, The FLAIR images were used with a field of view (FOV) of 192 × 220 mm² to 240 240 mm², voxel size of 0.43 × 0.43 mm² to 0.93 × 0.93 mm², slice thickness of 5 mm to 6 mm, echo-train length (ETL) of 4 to 27, repetition time (TR) of 7,468 ms to 9,000 ms, echo time (TE) of 78 ms to 128 ms, inversion time (TI) of 2,000 ms to 2,500 ms, and with or without fat saturation. The T2w sequence (2D fast spin echo) was used with an FOV of 192 × 220 mm² to 235 × 235 mm², voxel size of 0.28 × 0.28 mm² to 0.86 × 0.86 mm², slice thickness 2 mm to 6 mm, ETL of 16 to 27, TR of 4,290 to 9,000 ms and TE of 77 to 119 ms. The T1w sequence was used with an FOV of 192 × 220 mm² to 235 × 235 mm², voxel size of 0.43 × 0.43 mm² to 0.85 × 0.85 mm², slice thickness 4.8 mm to 10 mm, flip angle of 70° to 145°, TR of 250 to 714 ms and TE of 2.57 ms to 12.0 ms.

For the OASIS-3 dataset, The FLAIR images were used with a FOV of 183 × 210 mm² to 220 × 220 mm², voxel size of 0.42 × 0.42 mm² to 0.86 × 0.86 mm², slice thickness of 4 mm to 6 mm, ETL of 17 to 21, TR of 5,000 ms to 9,000 ms, TE of 76 ms to 94 ms, TI of 2,500 ms, and with or without fat saturation. The T2w sequence (2D fast spin echo) was used with an FOV of 176 × 256 × 144 mm³ to 256 × 256 × 256 mm³, voxel size of 1.0 × 1.0 × 4.0 mm³ to 0.51 × 0.51 × 0.5 mm³, ETL of 7 to 17, TR of 3,200 to 6,150 ms and TE of 86 to 116 ms. The T1w sequence was used with an FOV of 176 × 253 × 160 mm³ to 256 × 256 × 270 mm³, voxel size of 1.0 × 1.0 × 1.0 mm³ to 1.2 × 1.05 × 1.25 mm³, flip angle of 8° to 10°, TR of 7.3 to 9.7 ms and TE of 2.13 ms to 4.00 ms.

For the ADNI dataset [6], The T1w sequence (3D magnetization prepared rapid acquisition gradient echo) was used with FOV of 220 × 220 × 160 mm³ to 250 × 250 × 180 mm³, voxel size of 0.94 × 0.94 × 1.0 mm³ to 1.35 × 1.35 × 1.2 mm³, flip angle of 8° to 11°, TR of 6.5 to 11.04 ms and TE of 2.8 to 4.9 ms.

**Supplementary Information S3.** Details for the comparison methods

As comparison methods, compressed-sensing (CS) [7], U-Net [1], and variational network (VN) [8] were tested. For the structure of the U-Net, 32 feature channels and 5 max-pooling layers were applied. For the VN, the architecture consisted of four cascaded U-Nets, each configured with 24 feature channels and 5 max-pooling layers. A data consistency step, incorporating a learnable weighting factor, was applied after each block.

**Table S1.** Quantitative performance of the FLAIR prediction module on the public datasets.

| | | T1w → FLAIR | T2w → FLAIR | T1w, T2w → FLAIR |
|---|---|---|---|---|
| OASIS-3 | PSNR | 23.2 ± 3.7 | 25.8 ± 4.5 | 26.3 ± 4.0 |
| | SSIM | 0.880 ± 0.064 | 0.898 ± 0.060 | 0.912 ± 0.053 |
| fasrMRI | PSNR | 23.5 ± 3.9 | 24.7 ± 4.8 | 25.5 ± 4.6 |
| | SSIM | 0.887 ± 0.062 | 0.908 ± 0.051 | 0.914 ± 0.048 |

**Figure S1.** FLAIR prediction module results with modifying the scan parameters

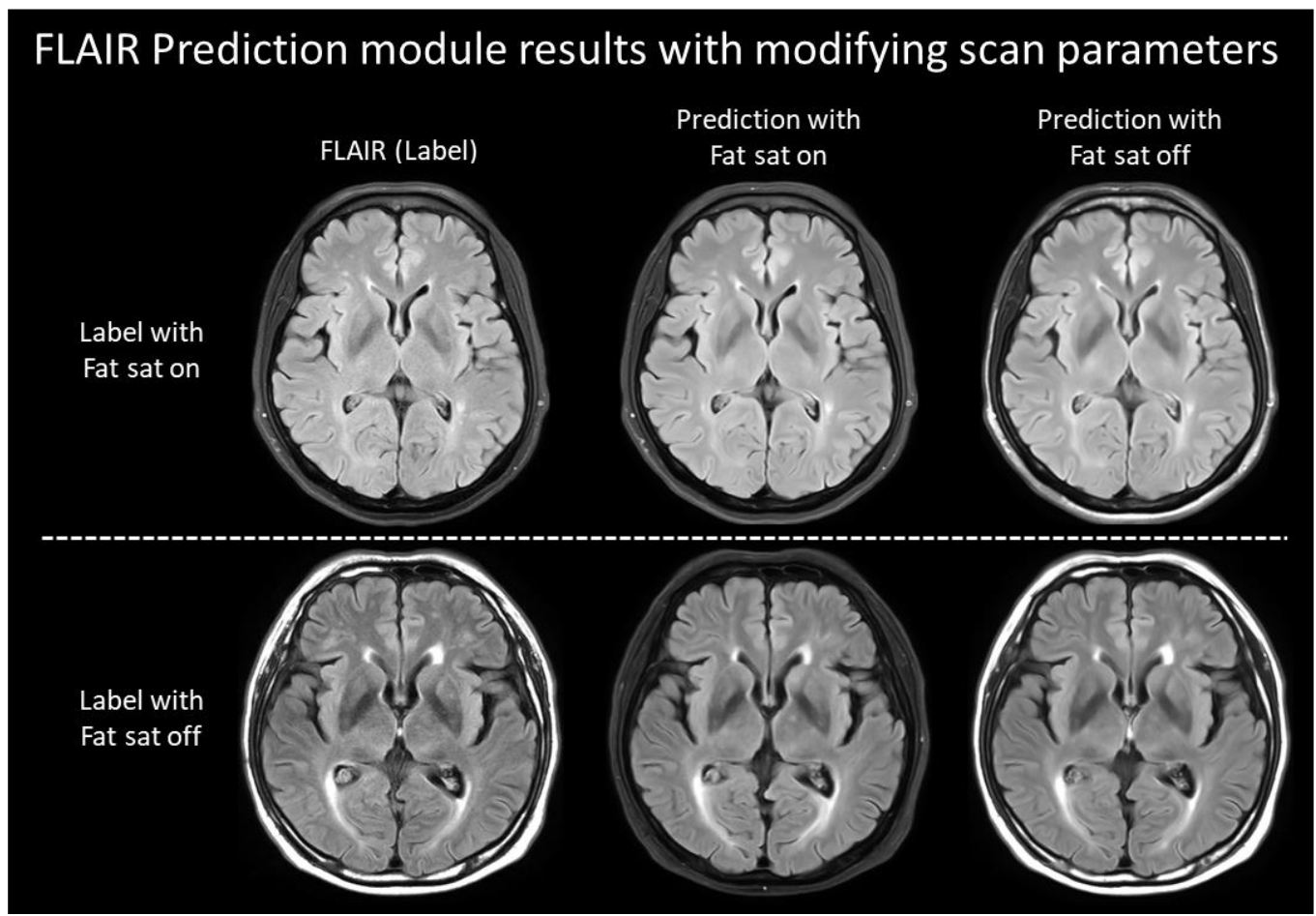

**Figure S1.** FLAIR prediction results with modifying scan parameters. The first row shows the experiments with the FLAIR image having a fat saturation scan parameter of 1. A FLAIR label image, a prediction image with fat saturation scan parameter of 1, and a prediction image with 0 is shown. The second row shows the experiments with the FLAIR image having a fat saturation scan parameter of 0. A FLAIR label image, a prediction image with fat saturation scan parameter of 1, and a prediction image with 0 is shown. In the skull region, significant changes were observed. The data imbalance (1,061 and 73 sessions for with and without fat saturation, respectively) results in a model that is highly optimized for the well-represented 'fat sat on' condition, leading to high-quality predictions for this state.

**Figure S2.** FLAIR Reconstruction results of acceleration ×4

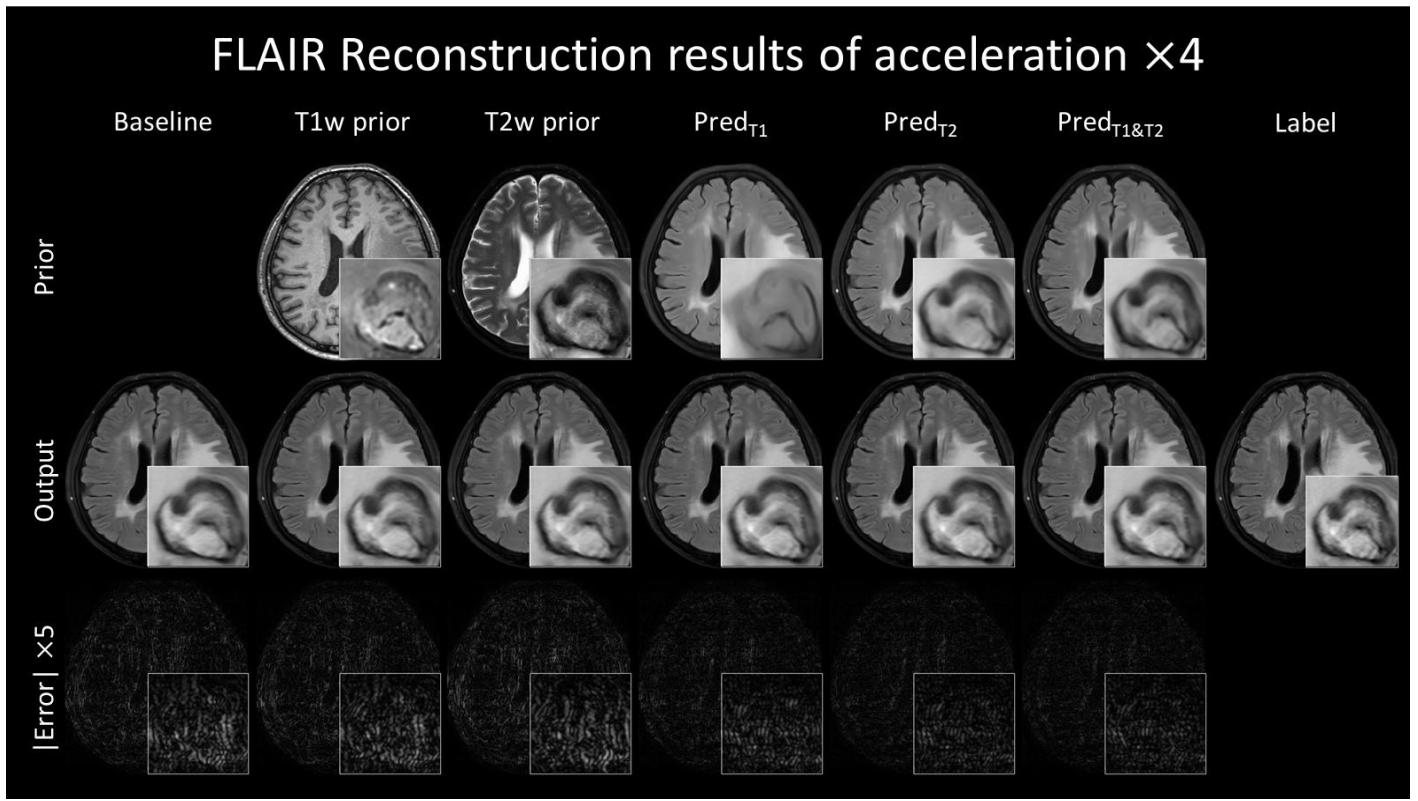

**Figure S2.** (a) Results of FLAIR reconstruction at an acceleration of ×4. The first row, the prior image for each method were shown. Reconstruction images for each comparison method were shown with labels in the final column (second row). The corresponding absolute error maps (×5) were shown in the last row. The proposed reconstruction method ($Pred_{T1}$, $Pred_{T2}$, and $Pred_{T1\&T2}$) show highly accurate results.

**Table S2.** Quantitative evaluation of the baseline network with other comparison methods.

|  |  | CS | U-Net | VN | Baseline |
|---|---|---|---|---|---|
| Acceleration ×4 | PSNR | 31.8 ± 1.9 | 32.1 ± 2.0 | 35.0 ± 1.9 | 35.1 ± 2.1 |
|  | SSIM | 0.916 ± 0.021 | 0.931 ± 0.013 | 0.961 ± 0.006 | 0.963 ± 0.008 |
| Acceleration ×8 | PSNR | 28.0 ± 2.0 | 28.4 ± 2.3 | 30.0 ± 2.0 | 30.2 ± 2.0 |
|  | SSIM | 0.853 ± 0.024 | 0.890 ± 0.014 | 0.909 ± 0.015 | 0.912 ± 0.013 |
| Acceleration ×12 | PSNR | 22.5 ± 2.2 | 23.6 ± 2.8 | 25.5 ± 2.2 | 25.9 ± 1.9 |
|  | SSIM | 0.806 ± 0.024 | 0.812 ± 0.021 | 0.837 ± 0.019 | 0.842 ± 0.016 |

**Table S3.** Quantitative evaluation of the reconstruction schemes on the fastMRI DICOM dataset.

|  |  | Baseline | T1 prior | T2 prior | $Pred_{T1}$ | $Pred_{T2}$ | $Pred_{T1\&T2}$ |
|---|---|---|---|---|---|---|---|
| Acceleration ×4 | PSNR | 34.8 ± 3.1 | 35.7 ± 3.5 | 35.8 ± 3.4 | 36.3 ± 3.1 | 36.7 ± 3.3 | 36.8 ± 3.3 |
|  | SSIM | 0.957 ± 0.006 | 0.965 ± 0.005 | 0.968 ± 0.005 | 0.975 ± 0.005 | 0.977 ± 0.004 | 0.978 ± 0.005 |
| Acceleration ×8 | PSNR | 30.5 ± 2.8 | 31.4 ± 3.2 | 31.6 ± 3.3 | 32.4 ± 3.0 | 33.0 ± 3.4 | 33.2 ± 3.3 |
|  | SSIM | 0.933 ± 0.012 | 0.940 ± 0.011 | 0.942 ± 0.009 | 0.948 ± 0.011 | 0.954 ± 0.008 | 0.955 ± 0.008 |
| Acceleration ×12 | PSNR | 26.3 ± 2.7 | 28.8 ± 3.3 | 29.0 ± 3.4 | 30.3 ± 2.9 | 31.1 ± 3.4 | 31.3 ± 3.3 |
|  | SSIM | 0.883 ± 0.018 | 0.915 ± 0.016 | 0.919 ± 0.013 | 0.928 ± 0.014 | 0.938 ± 0.013 | 0.940 ± 0.012 |

**Table S4.** Quantitative evaluation of the reconstruction schemes on the OASIS-3 dataset.

|  |  | Baseline | T1 prior | T2 prior | Pred$_{T1}$ | Pred$_{T2}$ | Pred$_{T1\&T2}$ |
|---|---|---|---|---|---|---|---|
| Acceleration ×4 | PSNR | 34.0 ± 1.2 | 35.1 ± 1.3 | 35.1 ± 1.3 | 35.2 ± 1.3 | 35.3 ± 1.3 | 35.2 ± 1.2 |
|  | SSIM | 0.959 ± 0.010 | 0.971 ± 0.008 | 0.972 ± 0.007 | 0.971 ± 0.007 | 0.972 ± 0.007 | 0.971 ± 0.006 |
| Acceleration ×8 | PSNR | 27.0 ± 1.1 | 27.8 ± 1.2 | 28.6 ± 1.1 | 30.2 ± 1.2 | 30.3 ± 1.2 | 30.4 ± 1.2 |
|  | SSIM | 0.902 ± 0.019 | 0.909 ± 0.019 | 0.918 ± 0.018 | 0.935 ± 0.015 | 0.934 ± 0.014 | 0.937 ± 0.015 |
| Acceleration ×12 | PSNR | 22.8 ± 1.0 | 24.8 ± 1.4 | 25.0 ± 1.1 | 27.7 ± 1.2 | 27.9 ± 1.1 | 28.0 ± 1.2 |
|  | SSIM | 0.843 ± 0.025 | 0.868 ± 0.030 | 0.870 ± 0.025 | 0.905 ± 0.021 | 0.907 ± 0.020 | 0.909 ± 0.021 |

**Table S5.** Quantitative evaluation on the fastMRI DICOM dataset trained with the internal dataset.

|  |  | Pred$_{T1}$ | Pred$_{T2}$ | Pred$_{T1\&T2}$ |
|---|---|---|---|---|
| Acceleration ×4 | PSNR | 36.2 ± 3.0 | 36.7 ± 3.2 | 36.7 ± 3.4 |
|  | SSIM | 0.973 ± 0.005 | 0.976 ± 0.005 | 0.977 ± 0.004 |
| Acceleration ×8 | PSNR | 32.3 ± 3.2 | 33.0 ± 3.3 | 33.1 ± 3.4 |
|  | SSIM | 0.946 ± 0.013 | 0.954 ± 0.008 | 0.954 ± 0.007 |
| Acceleration ×12 | PSNR | 30.1 ± 3.0 | 30.9 ± 3.5 | 31.1 ± 3.4 |
|  | SSIM | 0.926 ± 0.015 | 0.936 ± 0.012 | 0.938 ± 0.012 |

**Table S6.** Quantitative evaluation on the OASIS-3 dataset trained with the internal dataset.

|  |  | Pred$_{T1}$ | Pred$_{T2}$ | Pred$_{T1\&T2}$ |
|---|---|---|---|---|
| Acceleration ×4 | PSNR | 35.0 ± 1.3 | 35.2 ± 1.2 | 35.3 ± 1.2 |
|  | SSIM | 0.969 ± 0.007 | 0.971 ± 0.006 | 0.971 ± 0.006 |
| Acceleration ×8 | PSNR | 30.1 ± 1.4 | 30.2 ± 1.2 | 30.2 ± 1.3 |
|  | SSIM | 0.934 ± 0.015 | 0.934 ± 0.014 | 0.935 ± 0.015 |
| Acceleration ×12 | PSNR | 27.4 ± 1.3 | 27.7 ± 1.3 | 27.9 ± 1.2 |
|  | SSIM | 0.901 ± 0.021 | 0.905 ± 0.023 | 0.906 ± 0.022 |

**Table S7.** Quantitative evaluation on the multiple datasets trained with the combined dataset.

| Dataset |  | Internal dataset | | | OASIS-3 dataset | | | fastMRI DICOM dataset | | |
|---|---|---|---|---|---|---|---|---|---|---|
| Prior |  | Pred$_{T1}$ | Pred$_{T2}$ | Pred$_{T1\&T2}$ | Pred$_{T1}$ | Pred$_{T2}$ | Pred$_{T1\&T2}$ | Pred$_{T1}$ | Pred$_{T2}$ | Pred$_{T1\&T2}$ |
| Acceleration ×4 | PSNR | 36.2 ± 2.3 | 36.7 ± 2.2 | 36.8 ± 2.2 | 35.0 ± 1.5 | 35.3 ± 1.2 | 35.3 ± 1.3 | 36.5 ± 3.2 | 36.8 ± 3.3 | 36.8 ± 3.2 |
|  | SSIM | 0.972 ± 0.004 | 0.974 ± 0.007 | 0.975 ± 0.006 | 0.967 ± 0.009 | 0.971 ± 0.008 | 0.973 ± 0.007 | 0.976 ± 0.006 | 0.978 ± 0.005 | 0.978 ± 0.004 |
| Acceleration ×8 | PSNR | 31.1 ± 1.9 | 32.7 ± 2.2 | 32.9 ± 2.2 | 30.1 ± 1.2 | 30.4 ± 1.3 | 30.3 ± 1.4 | 32.3 ± 2.8 | 33.2 ± 3.3 | 33.3 ± 3.4 |
|  | SSIM | 0.924 ± 0.013 | 0.943 ± 0.011 | 0.940 ± 0.012 | 0.934 ± 0.014 | 0.936 ± 0.015 | 0.935 ± 0.016 | 0.945 ± 0.010 | 0.956 ± 0.008 | 0.956 ± 0.009 |
| Acceleration ×12 | PSNR | 23.0 ± 2.1 | 30.8 ± 2.1 | 30.9 ± 2.1 | 27.5 ± 1.3 | 27.7 ± 1.3 | 28.1 ± 1.1 | 30.4 ± 2.8 | 31.2 ± 3.3 | 31.2 ± 3.1 |
|  | SSIM | 0.893 ± 0.017 | 0.918 ± 0.015 | 0.921 ± 0.016 | 0.901 ± 0.024 | 0.904 ± 0.022 | 0.911 ± 0.020 | 0.929 ± 0.013 | 0.938 ± 0.013 | 0.939 ± 0.012 |

**Figure S3.** Longitudinal T1w reconstruction results of acceleration ×4

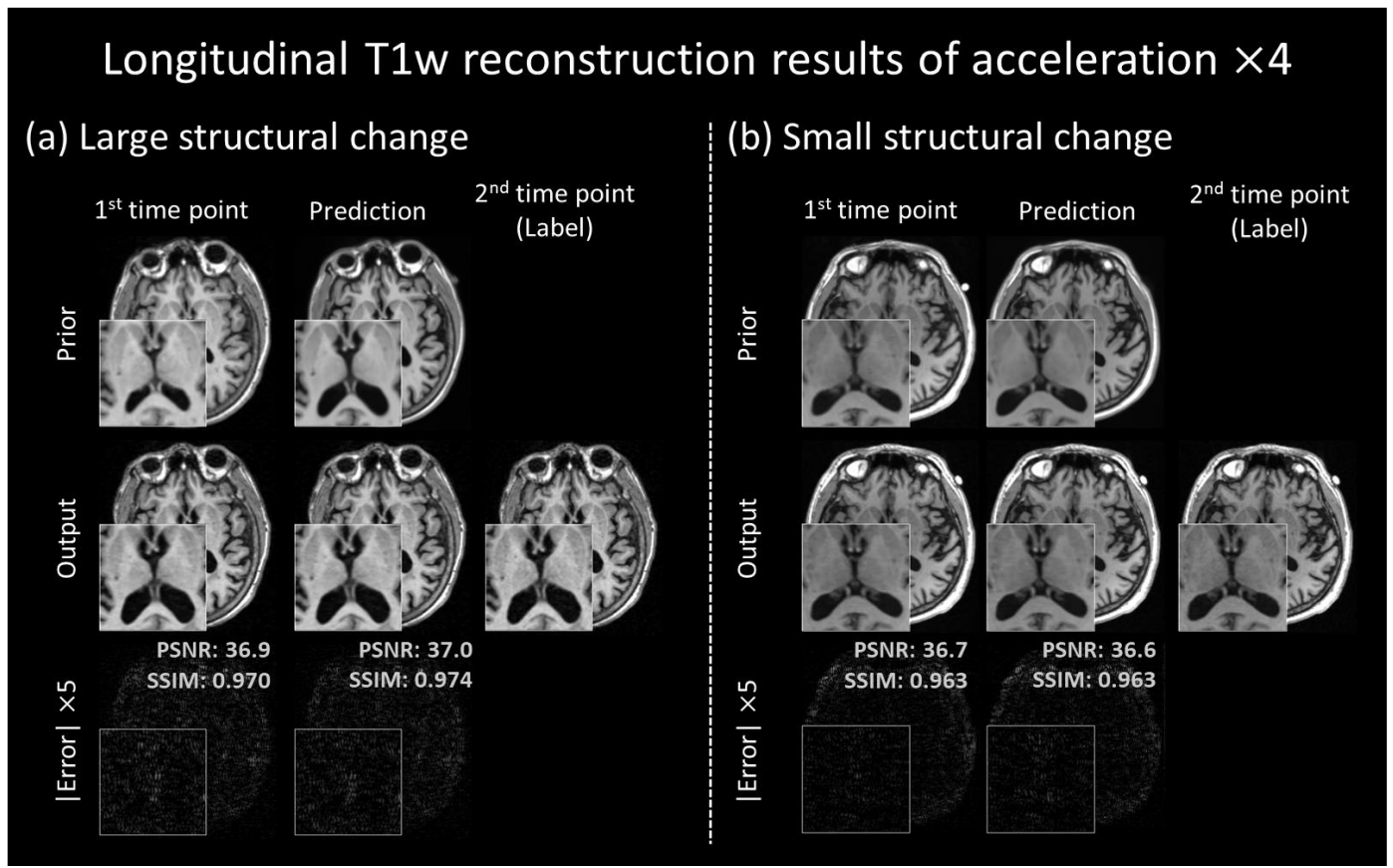

**Figure S3.** Longitudinal T1w reconstruction results on acceleration of ×4. Longitudinal pairs with prediction for large structural change (a) and small change (b). For the both pairs, reconstructions from the Longitudinal prior and our proposed method (Pred$_{long}$) are compared to the ground truth (label). Corresponding error maps and per-slice metrics are also shown.